\documentclass{article}

\usepackage[final, nonatbib]{neurips_2021_ml4ps}

\usepackage[utf8]{inputenc} 
\usepackage[T1]{fontenc}    
\usepackage{hyperref}       
\hypersetup{urlcolor=black,linkcolor=black,citecolor=black}
\usepackage{url}            
\usepackage{booktabs}       
\usepackage{amsfonts}       
\usepackage{nicefrac}       
\usepackage{microtype}      
\usepackage{xcolor}         
\usepackage{wrapfig}
\usepackage{adjustbox}
\usepackage[normalem]{ulem}
\usepackage{amsmath}
\usepackage{physics}
\usepackage{comment}

\usepackage{graphicx}
\usepackage[numbers,sort&compress]{natbib}
\title{
Laziness, Barren Plateau, and Noise in Machine Learning
}

\author{
    Junyu Liu \\
    Pritzker School of Molecular Engineering, The University of Chicago, Chicago, IL 60637, USA\\
    Chicago Quantum Exchange, Chicago, IL 60637, USA\\
    Kadanoff Center for Theoretical Physics, The University of Chicago, Chicago, IL 60637, USA\\
    qBraid Co., Harper Court 5235, Chicago, IL 60615, USA\\
    \texttt{junyuliu@uchicago.edu} \\
    \And
    Zexi Lin \\
    Pritzker School of Molecular Engineering, The University of Chicago, Chicago, IL 60637, USA\\
    \texttt{zexil@uchicago.edu} \\
    \And
    Liang Jiang \\
    Pritzker School of Molecular Engineering, The University of Chicago, Chicago, IL 60637, USA\\
    Chicago Quantum Exchange, Chicago, IL 60637, USA\\
    \texttt{liangjiang@uchicago.edu} 
}

\begin{document}

\maketitle

\begin{abstract}
We define \emph{laziness} to describe a large suppression of variational parameter updates for neural networks, classical or quantum. In the quantum case, the suppression is exponential in the number of qubits for randomized variational quantum circuits. We discuss the difference between laziness and \emph{barren plateau} in quantum machine learning created by quantum physicists in \cite{mcclean2018barren} for the flatness of the loss function landscape during gradient descent. We address a novel theoretical understanding of those two phenomena in light of the theory of neural tangent kernels. For noiseless quantum circuits, without the measurement noise, the loss function landscape is complicated in the overparametrized regime with a large number of trainable variational angles. Instead, around a random starting point in optimization, there are large numbers of local minima that are good enough and could minimize the mean square loss function, where we still have quantum laziness, but we do not have barren plateaus. However, the complicated landscape is not visible within a limited number of iterations, and low precision in quantum control and quantum sensing. Moreover, we look at the effect of noises during optimization by assuming intuitive noise models, and show that variational quantum algorithms are noise-resilient in the overparametrization regime. Our work precisely reformulates the quantum barren plateau statement towards a precision statement and justifies the statement in certain noise models, injects new hope toward near-term variational quantum algorithms, and provides theoretical connections toward classical machine learning. Our paper provides conceptual perspectives about quantum barren plateaus, together with discussions about the gradient descent dynamics in \cite{together}. 
\end{abstract}

\section{Barren plateau, laziness and noise}

Variational quantum circuits~\cite{peruzzo2014variational,yung2014transistor,mcclean2016theory,kandala2017hardware,cerezo2021variational, farhi2014quantum} can be used to optimize cost function measured on quantum computers. Specifically, these cost functions can be used for machine learning tasks~\cite{wittek2014quantum,wiebe2014quantum,biamonte2017quantum,schuld2019quantum,havlivcek2019supervised,liu2021rigorous,Liu:2021ohs,farhi2018classification}. In this case variational quantum circuits are addressed as quantum neural networks. 

However, a generically designed variational quantum ansatz may not be applicable to real problems. Specifically, a problem so-called \emph{barren plateau} has been widely discussed in the variational quantum algorithm community, which is believed to be one of the primary problems of quantum machine learning \cite{mcclean2018barren}. The argument is given as follows. A typical gradient descent algorithm will look like
\begin{align}\label{gds}
{\theta _\ell }(t + 1) - {\theta _\ell }(t) \equiv \delta {\theta _\mu } =  - \eta \frac{{\partial {\mathcal{L}}}}{{\partial {\theta _\ell }}}~,
\end{align}
where $\theta_\mu$ is the variational angle, and $t$ is referring the time step of gradient descent dynamics. $\eta$ is the learning rate, and $\mathcal{L}$ is the loss function. The observation \cite{mcclean2018barren} is that, if our variational ansatz is highly random, due to the $k$-design integral formula \cite{Roberts:2016hpo,Cotler:2017jue,Liu:2018hlr,Liu:2020sqb}, the derivative of the loss function is generically suppressed by the dimension of the Hilbert space $N$, and we might encounter a situation where the variation of the loss function during gradient descent is very small, namely $\delta \mathcal{L} \equiv \mathcal{L}(t+1)-\mathcal{L} (t)\ll 1$ for the step $t$. For instance, the second moment formula for Haar ensemble is 
\begin{align}\label{1design}
\int {dU{U_{ij}}U_{kl}^\dag }  = \frac{1}{N}{\delta _{il}}{\delta _{jk}}~.
\end{align}
Here $U$ is a unitary taken from a 1-design, and $\delta$ is the Kronecker delta and $i,j,k,l$ are matrix indexes. For higher moments random integrals \cite{Roberts:2016hpo,Cotler:2017jue,Liu:2018hlr,fukuda2019rtni,Liu:2020sqb}, the factor $\text{poly}(1/N)$ will appear. Thus, the difference between the variational angles during iterations will be suppressed by the dimension of the Hilbert space. The work \cite{mcclean2018barren} demonstrates this existence of the \emph{barren plateau} (the statement where $\delta \mathcal{L} \ll 1$) numerically and understands the result as a primary challenge of variational quantum circuits. It is often considered to be quantum analogs to the \emph{vanishing gradient problem}, but the nature is fundamentally different \cite{mohri2018foundations,roberts2021principles}. A further explanation is given in Appendix \ref{commentsclassical}.

Although the existence of the barren plateau is verified by numerous works \cite{cerezo2021cost,pesah2021absence,cerezo2021higher,arrasmith2021effect}, the theoretical understanding of the barren plateau problem is unclear. Moreover, the classical machine learning community has been successfully demonstrated its practical usage in science and business for years, and many successful classical neural network algorithms have been run for large scales. For example, \texttt{Generative Pre-trained Transformer-3} (\texttt{GPT-3}) from \texttt{OpenAI} \cite{brown2020language} has used 175 billion of training parameters, and it is one of the most successful natural language processing models up to date. Considering the standard LeCun initialization of weights $W$ with the normalization of the variance $\sigma_W^2$ \cite{mohri2018foundations,roberts2021principles,Liu:2021wqr}
\begin{align}\label{lecun}
\mathbb{E}({W_{ij}}W_{kl}^\dag ) = \frac{\sigma_W^2}{{{\rm{width}}}}{\delta _{ik}}{\delta _{jl}}~,
\end{align}
and its formal similarity to Equation \ref{1design}, we might imagine that similar issues will happen for classical neural networks too: they might be highly overparametrized in the large-width limit. Here, $\sigma_W$ is a number that is independent of the size of the neural networks, and we set the width of the neural network to be the same in each layer for simplicity. In fact, in Appendix \ref{commentsclassical}, we will show that in the classical large-width neural network, the barren plateau will also happen: the trainable weights do not run that much during gradient descent.

So, why classical overparametrized neural networks are supposed to be practical and good, but the barren plateaus of quantum neural networks are crucial challenges? In this paper, we define the primary theoretical argument towards the quantum barren plateau, the large suppression of the right hand side of Equation \ref{gds}, as \emph{laziness}. In the quantum context, the suppression is from the dimension of the Hilbert space, while in the classical case, the suppression is from the width of the classical neural networks. In a more precise language, laziness is referring to small $\delta \theta_\mu$, and barren plateau is referring to small $\delta \mathcal{L}$. 

Moreover, we will show that laziness may not imply the quantum barren plateau, from the perspective of overparametrization theory and representation learning theory through quantum neural tangent kernels (QNTKs) \cite{Liu:2021wqr,together}. In this paper, for quantum neural networks \emph{overparametrization} is referring to the fact where $ L \text{Tr}(O^2)/N^2 \approx \mathcal{O}(1)$, where $O$ is the operator we are optimizing, $L$ is the number of trainable angles, and $\eta$ is the learning rate as a constant. 

\begin{figure}[htp]
\centering
\includegraphics[width=1.0\textwidth]{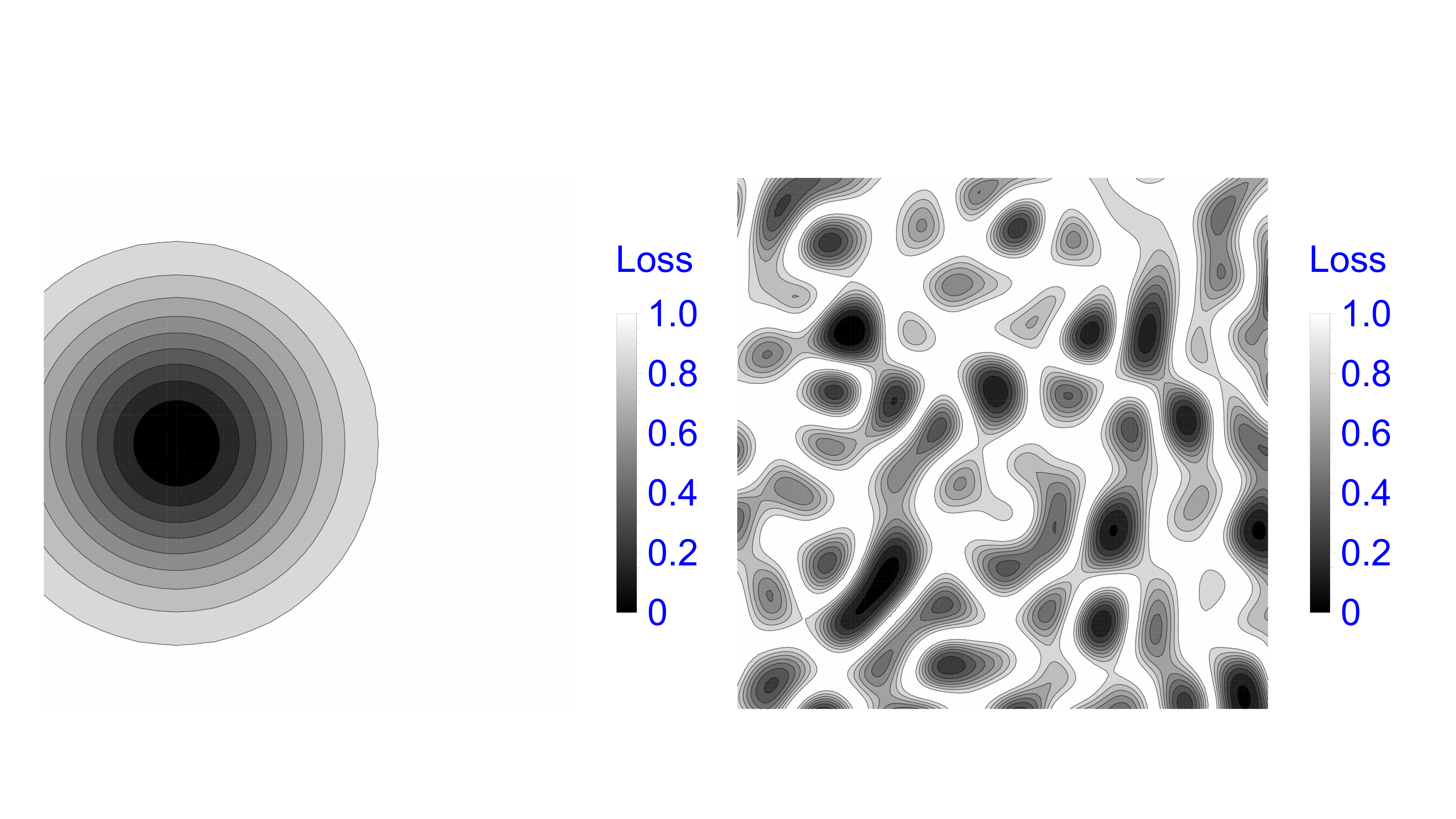}
\caption{Density plots of the loss function landscape comparing usual and overparametrized variational quantum circuits. We illustrate the landscape by color plots of the loss function for two variational angles. Left: the traditional understanding of \emph{barren plateaus} where we have the a single optimal point. Right: in the overparametrized case, the landscape is not barren, since for a random initial point, we get many good enough local optima that could minimize the loss function. Note that those plots are schematic since it is not possible to directly plot the loss function landscape in very high dimensions. In order to visualize it in $\mathcal{O}(1)$ numbers of iterations, one might have to have the number of trainable angles $L$ comparable to the dimension of the Hilbert space $N$.
}
\label{fig:barren}
\end{figure}

Defining quantum analogs of neural tangent kernels from their classical counterparts \cite{lee2017deep,jacot2018neural,lee2019wide,sohl2020infinite,yang2020feature,yaida2020non,arora2019exact,dyer2019asymptotics,halverson2021neural,roberts2021ai,roberts2021principles,summer}, we show that from the first-principle theoretical derivation, random (noiseless) quantum neural networks are still efficient to learn in the large-$L$ limit without barren plateaus, despite their laziness. In fact, although each trainable angle does not move much due to the small magnitude of the gradient, the combined effect of many of them on the loss function will still be significant. In addition, there exist good enough achievable local minima that minimize the training error. See Figure \ref{fig:barren} for an illustration. The requirements for making this to happen is especially when $L \text{Tr}(O^2)/N^2 \approx \mathcal{O}(1)$, and we have a small learning rate and the mean square loss function. In the case of large Hilbert space dimension without overparametrization, the exponential decay rate during gradient descent might be small, which may not make this phenomenon manifest in the polynomial training iterations. In practice, what we see is a very slow decay of loss functions. Interestingly, in this case quantum noises will not affect us significantly until exponential numbers of iterations. Thus, the averaged QNTK, $\bar{K}$, proportional to $\text{Tr}(O^2) L/N^2$, \emph{explains} the existence of the barren plateau in practice, with or without noises. On the other hand, in the overparametrization regime where $\eta L \text{Tr}(O^2)/N^2 \approx \mathcal{O}(1)$, the exponential decay of gradient descent process is visible.  

We note that the large-$L$ expansion is a quantum analog of the classical neural tangent kernel theory at large width. In fact, we will show in Section \ref{theory} that we have similar large-width expansion comparing the classical theory, where in our model, \emph{classical width} corresponds to $L$. The dimension of the Hilbert space plays an important role in the calculation. Moreover, the correspondence between quantum and classical neural networks might be explained by some physical heuristics, from the duality between matrix models and quantum field theories. See Appendix \ref{stringtheory} for a brief discussion.

Moreover, we need to point out that laziness is intrinsically still a precision problem. More precisely, it could be primarily from quantum measurement and quantum control, since the size of classical devices could scale as $\log (1/\epsilon)$ for given precision $\epsilon$, while variational quantum circuits cannot, due to the measurement error and the limitation of quantum control \cite{mcclean2018barren}. Thus, it naturally motivates us to think about how to include the effect of noise in the gradient descent calculation. In our work, we introduce a simple and intuitive noise model by adding random variables in the gradient descent dynamics. We show that in the overparametrization regime, our variational quantum algorithms are noise-resilient. More precisely, we find that the residual training error scales as 
\begin{align}
{\varepsilon ^2}(t) \approx {(1 - \eta K)^{2t}}\left( {{\varepsilon ^2}(0) - \frac{{\sigma _\theta ^2}}{{\eta (2 - \eta K)}}} \right) + \frac{{\sigma _\theta ^2}}{{\eta (2 - \eta K)}}~,
\end{align}
with the neural tangent kernel $K$ and the standard deviation of the noise introduced in the variational angles $\sigma_\theta$. Thus, in the late time, we get
\begin{align}
\mathcal{L}(\infty ) = \frac{1}{2}{\varepsilon ^2}(\infty ) \approx \frac{{\sigma _\theta ^2}}{{2\eta (2 - \eta K)}}~.
\end{align}
In the late time, we have 
\begin{align}
\mathcal{L}(\infty ) = \frac{1}{2}{\varepsilon ^2}(\infty ) \approx \frac{{\sigma _\theta ^2}}{{2\eta (2 - \eta K)}}~.
\end{align}
Thus, in the overparametrized regime, we could set $\eta K \approx \mathcal{O}(1)$, so schematically,
\begin{align}
\mathcal{L}(\infty ) \approx \mathcal{O}(\frac{{\sigma _\theta ^2}}{\eta })~,
\end{align}
indicating that we could get good predictions at the end as long as we sufficiently control the noises. 

We will give more details in the following sections.

\section{The loss function landscape and the QNTK theory}\label{theory}
We begin by considering a variational quantum circuit ansatz, on a Hilbert space of size $N$ with $\log_2 N$ qubits, as follows,
\begin{align}
   U(\theta ) = \left( {\prod\limits_{\ell  = 1}^L {{W_\ell }} \exp \left( {i{\theta _\ell }{X_\ell }} \right)} \right) \equiv \left( {\prod\limits_{\ell  = 1}^L {{W_\ell }} U_{\ell}}\right),
\end{align}
with some trainable angles $\theta_\ell$, constant unitary operators  $W_\ell$, and Pauli operators $X_\ell$. Following~\cite{Liu:2021wqr}, we consider the mean square loss function
\begin{align}\label{opt_problem}
{\cal L}(\theta ) = \frac{1}{2}{\left( {\left\langle {{\Psi _0}\left| {{U^\dag }(\theta )OU(\theta )} \right|{\Psi _0}} \right\rangle  - {O_0}} \right)^2} \equiv \frac{1}{2}{\varepsilon ^2}~,
\end{align}
and train the expectation value $\left\langle {{\Psi _0}\left| {{U^\dag }(\theta )OU(\theta )} \right|{\Psi _0}} \right\rangle $ on an initial state $\ket{\Psi_0}$ towards a value $O_0$. We define the residual training error $\varepsilon  = \left\langle {{\Psi _0}\left| {{U^\dag }(\theta )OU(\theta )} \right|{\Psi _0}} \right\rangle  - {O_0}$. We use the gradient descent algorithm Equation \ref{gds} with the learn ing rate $\eta$ and an initial variational angle $\theta(0)$. We look now at the difference of the residual training error
\begin{align}
\delta \varepsilon  \equiv \varepsilon (t + 1) - \varepsilon (t)~.
\end{align}
When the learning rate of Equation \ref{gds} $\eta$ is small, we can perform a Taylor expansion,
\begin{align}\label{taylor}
\delta \varepsilon  \approx \sum\limits_\ell  {\frac{{\partial \varepsilon }}{{\partial {\theta _\ell }}}} \delta {\theta _\ell } =  - \eta \sum\limits_\ell  {\frac{{\partial \varepsilon }}{{\partial {\theta _\ell }}}} \frac{{\partial \varepsilon }}{{\partial {\theta _\ell }}}\varepsilon  =  - \eta K\varepsilon ~.
\end{align}
The quantity $K$ here is called the Quantum Neural Tangent Kernel (QNTK) \cite{Liu:2021wqr}, $K = \sum\limits_\ell  {\frac{{\partial \varepsilon }}{{\partial {\theta _\ell }}}} \frac{{\partial \varepsilon }}{{\partial {\theta _\ell }}}$. Note that in a general supervised learning setup where one has a labeled dataset instead of just one expected value $O_0$, $K$ is a positive-semidefinite and symmetric matrix instead of a non-negative number. Here we focus on the optimization problem Equation \ref{opt_problem}: this example will demonstrate the validity of our theory, that can be readily generalized to a full supervised quantum machine learning setup.

A {\it frozen} QNTK will remain constant during a gradient descent flow  will lead to gradient flow equations which can be solved exactly~\cite{Liu:2021wqr}, showing that the error will decay exponentially at the gradient descent iteration $t$ as
\begin{align}\label{varep}
\varepsilon (t) = {(1 - \eta K)^t}\varepsilon (0)~.
\end{align}
For sufficient random variational ans\"{a}tze, we could compute the value of $K$ based on the same assumption of the barren plateau problem \cite{mcclean2018barren}. After computing 2-design random average $\mathbb{E}$ (see \cite{together} for more details)
\begin{align}
\mathbb{E}(O) = \int_{U \in {\text{2-design}}} {dUO(U)}~, 
\end{align}
More precisely, we define
\begin{align}
&{U_{ - ,\ell }} \equiv \prod\limits_{\ell ' = 1}^{\ell  - 1} {{W_{\ell '}}} {U_{\ell '}},{U_{ + ,\ell }} \equiv \prod\limits_{\ell ' = \ell  + 1}^L {{W_{\ell '}}} {U_{\ell '}},\nonumber\\
&{V_{ - ,\ell }} = {U_{ - ,\ell }}{W_\ell }{U_\ell },{V_{ + ,\ell }} = {U_{ + ,\ell }}~.
\end{align}
And we assume that ${V_{ - ,\ell }}$ and ${V_{ + ,\ell }}$ form 2-designs independently in all $\ell$s. We get the following expression of the averaged QNTK,
\begin{align}\label{fullqntke}
&\bar K = {\mathbb{E}}(K) = L\left( {N{\mathop{\rm Tr}\nolimits} \left( {{O^2}} \right) - {{{\mathop{\rm Tr}\nolimits} }^2}(O)} \right)\frac{2}{{N + 1}}\left( {\frac{1}{{{N^2} - 1}}} \right)  \nonumber\\
&\approx \frac{{2L{\mathop{\rm Tr}\nolimits} \left( {{O^2}} \right)}}{{{N^2}}} ~.
\end{align}
This simple equation combined with Equation \ref{varep} reveals how, on average, the residual training error of a gradient descent dynamics will decay exponentially. Moreover, one should also check the standard deviation $\Delta K$. If $\Delta K \ll \bar{K}$, we get a distribution of $K$ which is concentrated at $\bar{K}$. In fact, one could show that from $k$-design assumptions,
\begin{align}
\Delta K \approx \frac{{\sqrt L }}{{{N^2}}}\sqrt {\left( {8{\rm{Tr}}{^2}\left( {{O^2}} \right) + 12{\rm{Tr}}\left( {{O^4}} \right)} \right)} ~.
\end{align}
Thus, we have $\Delta K /\bar{K} =\mathcal{O}(1/\sqrt{L})$. In the limit where $L\gg 1$, the neural tangent kernel is concentrated around a fixed value $\bar{K}$. A more precise constraint will also include a time-dependent statement including the perturbations of higher-order Taylor expansion of the residual training error, which is characterized by the so-called quantum meta-kernel or dQNTK. See Appendix \ref{concenreview} for more details.

\section{Precision and noise}\label{deep}
Now we give some physical interpretations about Equation \ref{fullqntke}. We see in Section \ref{theory} that the theory should work in the regime where $L \gg 1$, and also the overparametrization regime where $\eta K\approx \mathcal{O}(1)$. From Equation \ref{varep}, we know that $\bar{K}$ would serve as an exponent of exponential decay: the larger $\bar{K}$ is, the faster the algorithm will converge. This qualitative description has been formulated in \cite{Liu:2021wqr}, with numerical evidence in \cite{shirai2021quantum} around the same time. 

Moreover, a statement about precision could be made by combining Equation \ref{varep} and Equation \ref{fullqntke}. We have
\begin{align}
\log \frac{1}{{{\varepsilon _r}}} \approx  - T\log (1 - \eta \bar{K}) \approx \eta \bar{K}T~.
\end{align}
Here, $T$ is the total training steps, and $\varepsilon_r$ is the relative residual training error around the end of training $\varepsilon_r = \varepsilon(T)/\varepsilon(0)$. The relative error $\varepsilon_r$ could be as small as the precision of the quantum device. Using Equation \ref{fullqntke}, we get
\begin{align}\label{depth}
\log \frac{1}{{{\varepsilon _r}}} \approx \frac{{2\eta L{\rm{Tr}}\left( {{O^2}} \right)T}}{{{N^2}}} ~.
\end{align}
Equation \ref{depth} makes the barren plateau problem manifestly as a precision problem. If we want to see the convergence within $T\approx \mathcal{O}(1)$, we want $\eta \bar{K} \approx 1$. The smaller $\bar{K}$ is, the smaller decaying exponent we have, and more likely we will experience a barren plateau in practice. Otherwise, there will be good enough local optima around the small random fluctuations of variational angles. The more overparametrized the quantum neural networks are, the faster convergence they could have. In this case, we do not have a barren plateau if we assume that we do not have the measurement noise and the quantum hardware noise, although we have laziness.  

Originally, a relation between the barren plateau problem and the precision has also been stated in \cite{mcclean2018barren}, while we make it more clear by showing that the barren plateau is not algorithmic. In fact, in Appendix \ref{commentsclassical}, we show that classical overparametrized neural networks have laziness as well. Many useful, practical machine learning algorithms have to be in this case \cite{roberts2021principles}. Thus, variational quantum algorithms here have no algorithmic issue, and the origin of the problem comes from measurement and control (see also \cite{wang2021noise}).

Let us take a look at Equation \ref{gds} again. To implement variational algorithms, we need to perform measurements to evaluate the loss function or its derivatives (involving quantum measurements), and update the trainable angles through Equation \ref{gds} (involving quantum control). On the measurement side, classical computations could handle the precision-$\epsilon$ computation with the resource scaling as $\log 1/\epsilon$, while measurement errors will be produced in the quantum setup, making the scaling $1/\epsilon ^\alpha$ for positive $\alpha$ \cite{mcclean2018barren}. There is no known way to date to avoid it because of limitations of metrology \cite{knill2007optimal}. On the control side, it is also challenging to update the variational angles with exponential precision. In a sense, our theory makes the statement from \cite{mcclean2018barren} more precise. 

The discussion naturally motivates us to introduce the noise model. Heuristically, we will expect that during the gradient descent process, the effective noise term will also be exponentially decaying because of the original recurrence relation and its solution. To verify this, we could add a random fluctuation term $\Delta \theta_\ell$ to model the uncertainty of measuring the expectation value. One could also assume that the random variable $\Delta \theta_\ell$ is Markovian. Namely, it is independent for the time step $t$. Moreover, we assume that $\Delta \theta _\ell $s are distributed with Gaussian distributions $\mathcal{N}(0,\sigma^2_\theta)$. Note that $\sigma_\theta$ could come from the measurement noise during estimations of quantum observables used for the gradient descent, which scales as $1/\sqrt{n}$, where $n$ is the number of measurements. And the Gaussian assumptions come from the central limit theorem in the large-$n$ limit. Furthermore, $\sigma_\theta$ could also come from the hardware noises. On the other hand, the physical implementation of rotation angle will also have limited precision. One could note that robust quantum control techniques can suppress errors of rotation angles to higher orders, see \cite{vandersypen2005nmr}. 

Thus, one could show that the residual training error has the recursion relation in the linear order of the Taylor expansion,
\begin{align}
\delta \varepsilon =  - \eta \varepsilon K + \sum\limits_\ell  {\frac{{\partial \varepsilon }}{{\partial {\theta _\ell }}}\Delta \theta _\ell } ~.
\end{align}
Now, let us assume that $K$ is still a constant, $K\approx \bar{K}$. Since $\Delta \theta_\ell \sim \mathcal{N}(0,\sigma^2_\theta)$, we get
\begin{align}
\sum\limits_\ell  {\frac{{\partial \varepsilon }}{{\partial {\theta _\ell }}}\Delta {\theta _\ell }} \sim \mathcal{N}(0,K\sigma _\theta ^2)~.
\end{align}
Including the noise term into the recursion relation, one could show that averaging over the random distribution of the noise, we have 
\begin{align}
\overline {{\varepsilon ^2}} (t)= {(1 - \eta K)^{2t}}\left( {{\varepsilon ^2}(0) - \frac{{\sigma _\theta ^2}}{{\eta (2 - \eta K)}}} \right) + \frac{{\sigma _\theta ^2}}{{\eta (2 - \eta K)}}~.
\end{align}
Note that the first term is decaying when the time $t$ is increasing. At the late time, we have
\begin{align}
\overline {{\varepsilon ^2}} (\infty ) = \frac{{\sigma _\theta ^2}}{{\eta (2 - \eta K)}} \approx \mathcal{O}(\frac{{\sigma _\theta ^2}}{\eta })~,
\end{align}
where we assume the overparametrization $\eta K \approx \mathcal{O}(1)$~. Thus, at the late time, the loss function will arrive at a constant plateau at $\mathcal{O}({{\sigma _\theta ^2}}/{\eta })$. One could improve $\sigma_\theta$ to make the constant plateau controllable and do not increase significantly with $N$, indicating that our algorithm could be noise-resilient. See Appendix \ref{noise} for a more detailed discussion, and see Figure \ref{fig:barren} for an illustration. Some numerical results are also obtained in Figure \ref{fignew} and Figure \ref{sigmathetavseta}.

\section{Conclusion and outlook}
In this paper, we point out that for variational circuits with sufficiently large numbers of trainable angles, the gradient descent dynamics could still be efficiently performed, despite the existence of the exponential suppression of the variational angle updates (laziness). We point out that laziness is not uniquely happening in quantum machine learning, but also for overparametrized classical neural networks with large widths. The efficiency of large-width neural networks is justified by the neural tangent kernel theory, so do their quantum counterparts. A solid and simple theory has been established based on the above ideas, and the relation between the number of training steps, the quantum device error, the trainable depth, the dimension of the Hilbert space, and the norm of operators appearing in the loss function has been explicitly derived. Moreover, we have justified that for simple and natural noise models, we could make the variational quantum circuits noise-resilient in the overparametrized regime, with solid theoretical and numerical evidence. 

Our results also indicate a more well-defined path to designing quantum neural networks from the first principle. If we are sampling unitary operators uniformly in the whole unitary group, it is hard to avoid polynomial factors of $N$, the dimension of the Hilbert space, into the expression of the number of iterations in order to obtain the visible laziness (see parallel efforts in \cite{abedi2022quantum,You:2022tsl}). One idea is to reduce the space of searching, and reduce the space of variational circuits to some subspaces, where people observe some evidence for setups in quantum convolutional neural networks \cite{cong2019quantum,pesah2021absence} and local loss function \cite{cerezo2021cost}, and the barren plateau phenomena are less drastic in those cases. However, since the subspace we are searching is reduced, the decreased expressibility will lead to a lower performance for the final convergence of the loss function on the training set \cite{abedi2022quantum}: around the end of the training, drastic corrections towards fixed neural tangent kernels will stop the exponential decay, and we get a local minimum which may not be good enough. The design of variational circuits will be a trade-off between barren plateaus and performance \cite{larocca2021theory}, which could be manifest in the presence of laziness. Despite generalizations to full learning setups with multiple output dimensions, other interesting directions include detailed discussions about the quantum noise in the real machines during quantum representation learning to understand how the noise will affect laziness and the barren plateau, a justification of our theory with large-scale classical and quantum simulation, and possible theoretical understandings beyond the limit $L\gg 1$. We look forward to further analysis and research along our path. 

\emph{Note added:} When the paper is finished, we notice that another nice independent paper \cite{Anschuetz:2022wvo} appears in the arxiv, which has very similar conclusion to our results. 


\begin{ack}

We thank Jens Eisert, Keisuke Fujii, Isaac Kim, Risi Kondor, Kenji Kubo, Antonio Mezzacapo, Kosuke Mitarai, Khadijeh Najafi, Sam Pallister, John Preskill, Dan A. Roberts,  Norihito Shira, Eva Silverstein, Francesco Tacchino, Shengtao Wang, Xiaodi Wu, Yi-Zhuang You, Han Zheng, and Quntao Zhuang for useful discussions.

We acknowledge support from the ARO (W911NF-18-1-0020, W911NF-18-1-0212), ARO MURI (W911NF-16-1-0349, W911NF-21-1-0325), AFOSR MURI (FA9550-19-1-0399, FA9550-21-1-0209), AFRL (FA8649-21-P-0781), DoE Q-NEXT, NSF (OMA-1936118, EEC-1941583, OMA-2137642), NTT Research, and the Packard Foundation (2020-71479).

\end{ack}

\clearpage

\pagebreak

\appendix

\vspace{0.5in}

\begin{center}
	{\Large \bf Appendix}
\end{center}

\section{Comments on the barren plateau in the \emph{classical} machine learning}\label{commentsclassical}
Now we consider a classical neural network, the MLP model (see \cite{roberts2021principles}). The definition is
\begin{align}
&z_i^{(1)}\left( {{x_\alpha }} \right) \equiv b_i^{(1)} + \sum\limits_{j = 1}^{{n_0}} {W_{ij}^{(1)}} {x_{j;\alpha }},\nonumber\\
&{\rm{for}}\quad i = 1, \ldots ,{n_1},\nonumber\\
&z_i^{(\ell  + 1)}\left( {{x_\alpha }} \right) \equiv b_i^{(\ell  + 1)} + \sum\limits_{j = 1}^{{n_\ell }} {W_{ij}^{(\ell  + 1)}} \sigma \left( {z_j^{(\ell )}\left( {{x_\alpha }} \right)} \right),\nonumber\\
&{\rm{ for }}\quad i = 1, \ldots ,{n_{\ell  + 1}};\ell  = 1, \ldots ,L - 1.
\end{align}
Here, $\sigma$ is a non-linear activation function, and we have widths $n_{1,2,\cdots,L}$ in layers $\ell=1,2,\cdots L$. The input dimension is $n_0$ and the output dimension is $n_L$. Weights and biases at layer $\ell$ are denoted as $W^{(\ell)}$ and $b^{(\ell)}$. $z^{(\ell)}$ is called the \emph{preactivation}. $x_{j,\alpha}$ will denote the data where $j$ is the vector index, and $\alpha$ is the data sample index. At the beginning, we initialize the neural network by
\begin{align}\label{lecunformula}
&{\mathbb{E}}\left[ {b_{{i_1}}^{(\ell )}b_{{i_2}}^{(\ell )}} \right] = {\delta _{{i_1}{i_2}}}C_b^{(\ell )}~,\nonumber\\
&{\mathbb{E}}\left[ {W_{{i_1}{j_1}}^{(\ell )}W_{{i_2}{j_2}}^{(\ell )}} \right] = {\delta _{{i_1}{i_2}}}{\delta _{{j_1}{j_2}}}\frac{{C_W^{(\ell )}}}{{{n_{\ell  - 1}}}}~.
\end{align}
Here, $C_b$ and $C_W$ will set the variance of biases and weights (we use the notation $C_W=\sigma_W^2$ in the main text). And we train the neural networks by gradient descent algorithms. We could consider the simplest version of the gradient descent algorithm, 
\begin{align}
{\theta _\mu }(t + 1) = {\theta _\mu }(t) - {\left. {\eta \frac{{d{{\cal L}_{\cal A}}}}{{d{\theta _\mu }}}} \right|_{\theta (t)}}~.
\end{align}
The loss function is
\begin{align}
{{\cal L}_{\cal A}} \equiv \frac{1}{2}\sum\limits_{i,\tilde \alpha  \in {\cal A}} {{{\left( {{z_i}\left( {{x_{\tilde \alpha }};\theta } \right) - {y_{i,\tilde \alpha }}} \right)}^2}}  = \frac{1}{2}\sum\limits_{i,\tilde \alpha  \in {\cal A}} {\varepsilon _{i,\tilde \alpha }^2} ~,
\end{align}
where $\tilde{\alpha}\in \mathcal{A}$ form a training set $\mathcal{A}$, and we have a supervised learning task with the data label $y$. $z_i$ is the final prediction from the MLP model, $z_{i}^{(L)}$, $\eta$ is the training rate. $\theta_\mu$ is a vector combining all $W$s and $b$s. $\varepsilon$ here is the residual training error,
\begin{align}
{\varepsilon _{i,\tilde \alpha }} = {z_i}\left( {{x_{\tilde \alpha }}} \right) - {y_{i,\tilde \alpha }}~.
\end{align}

\subsection{The fundamental difference between barren plateau and vanishing gradient}
Firstly, we wish to comment on the fact that there is a fundamental difference between the barren plateau problem and the vanishing gradient problem. 

The vanishing gradient problem is claimed to be a challenge of machine learning algorithms, where the gradient is vanishing for some neural network constructions, and it will be challenging to train the network \cite{hochreiter1998vanishing,hochreiter2001gradient}. A standard and traditional explanation of the vanishing gradient problem is due to multiplicatively large number of layers in a deep neural network. The loss will have exponential behavior against some multiplicative factors during gradient descent, which will cause either exploding or vanishing of the loss function if there is no fine tuning. A resolution of the vanishing gradient problem is associated with the idea of \emph{He initialization} or \emph{Kaiming initialization}, which fine-tunes the neural network towards its critical point \cite{he2015delving} (see also \cite{roberts2021principles}). 

The \emph{barren plateau problem} is a term invented from the quantum community since \cite{mcclean2018barren}. As far as we know, there is no such term in classical machine learning instead of geography. The theoretical argument from the barren plateau problem is the following, where we define the argument as \emph{laziness}. If we consider the gradient descent process of the variational angles,
\begin{align}
{\theta _\mu }(t + 1) = {\theta _\mu }(t) - {\left. {\eta \frac{{d{{\cal L}_{\cal A}}}}{{d{\theta _\mu }}}} \right|_{\theta (t)}}~.
\end{align}
and if we make a sufficiently random variational ansatz, the factor $\text{poly}(\dim \mathcal{H})$ where $\dim \mathcal{H}$ is the dimension of the Hilbert space, will appear in the formula of $d\mathcal{L}_A/d \theta_\mu$. Thus, the change of the variational angle will always suppressed by the dimension of the Hilbert space. A simple example of the Haar random factor $\text{poly}(\dim \mathcal{H})$ will be the integration formula over a 2-design,
\begin{align}\label{designformula}
\int {dU{U_{ij}}U_{kl}^\dag }  = \frac{{{\delta _{il}}{\delta _{jk}}}}{{\dim {\cal H}}}~,
\end{align}
where the matrix $U$ forms a 2-design. The higher $k$ is in a $k$-design, the higher factor of $\dim \mathcal{H}$ will appear if we consider higher moments of $U$. Thus, one claim that the variational angles almost cannot run in the randomized variational quantum architectures. 

We could notice that the argument of the barren plateau problem using laziness is fundamentally different from the vanishing gradient problem: the vanishing gradient problem is \emph{dynamical} when going to deeper and deeper neural networks, while the laziness is \emph{static} and appears everywhere. Thus they are two intrinsically different problems. Moreover, from the similarity between the 2-design integral formula \ref{designformula} and the LeCun parametrization \ref{lecunformula}, we could expect that the large-width neural networks will have similar behaviors: their weights and biases will also almost not run. Considering that classical overparametrized neural networks are proven to be practically useful (see, for instance, a comparison \cite{golubeva2020wider}), and large-scale neural networks could be implemented commonly nowadays, laziness may not always be bad in the actual machine learning tasks.

\subsection{Classical large-width neural network has laziness as well}
Now we prove that in the above setup, the large-width classical neural network will also have laziness. We have
\begin{align}
&\frac{{d{{\cal L}_{\cal A}}}}{{d{\theta _\mu }}} = \sum\limits_{i,\tilde \alpha } {{\varepsilon _{i,\tilde \alpha }}\frac{{d{\varepsilon _{i,\tilde \alpha }}}}{{d{\theta _\mu }}}}  = \sum\limits_{i,\tilde \alpha } {{\varepsilon _{i,\tilde \alpha }}\frac{{d{z_{i,\tilde \alpha }}}}{{d{\theta _\mu }}}}  \nonumber\\
&=\sum\limits_{i,\tilde \alpha } {{y_{i,\tilde \alpha }}\frac{{d{z_{i,\tilde \alpha }}}}{{d{\theta _\mu }}}}  + \sum\limits_{i,\tilde \alpha } {{z_{i,\tilde \alpha }}\frac{{d{z_{i,\tilde \alpha }}}}{{d{\theta _\mu }}}} ~.
\end{align}
We wish to represent the derivatives over $W$ and $b$ by the derivatives of early-layer preactivation $z^{(\ell)}$,
\begin{align}\label{basicMLP}
&\frac{{dz_{i;\alpha }^{(L)}}}{{db_j^{(\ell )}}} = \frac{{dz_{i;\alpha }^{(L)}}}{{dz_{j;\alpha }^{(\ell )}}}~,\nonumber\\
&\frac{{dz_{i;\alpha }^{(L)}}}{{dW_{jk}^{(\ell )}}} = \sum\limits_m {\frac{{dz_{i;\alpha }^{(L)}}}{{dz_{m;\alpha }^{(\ell )}}}} \frac{{dz_{m;\alpha }^{(\ell )}}}{{dW_{jk}^{(\ell )}}} = \frac{{dz_{i;\alpha }^{(L)}}}{{dz_{j;\alpha }^{(\ell )}}}\sigma _{k;\alpha }^{(\ell  - 1)}~.
\end{align}
Here, $\sigma^{(\ell)}$ is a short-hand notation of $\sigma(z^{(\ell)})$, and we introduce $\sigma_{j;\alpha}^{(\ell)}$ as $\sigma (z_{j;\alpha}^{(\ell)})$. Finally, we have,
\begin{align}
&\frac{{dz_{i;\alpha }^{(L)}}}{{dz_{j;\alpha }^{(\ell )}}} = \sum\limits_{k = 1}^{{n_{\ell  + 1}}} {\frac{{dz_{i;\alpha }^{(L)}}}{{dz_{k;\alpha }^{(\ell  + 1)}}}} \frac{{dz_{k;\alpha }^{(\ell  + 1)}}}{{dz_{j;\alpha }^{(\ell )}}} = \sum\limits_{k = 1}^{{n_{\ell  + 1}}} {\frac{{dz_{i;\alpha }^{(L)}}}{{dz_{k;\alpha }^{(\ell  + 1)}}}} W_{kj}^{(\ell  + 1)}\sigma^{(\ell)} _{j;\alpha }{'}\nonumber\\
&{\rm{ for }}\quad \ell  < L ~,\nonumber\\
&\frac{{dz_{i;\alpha }^{(L)}}}{{dz_{j;\alpha }^{(L)}}} = {\delta _{ij}}~.
\end{align}
This is a back-propagation iterative formula, giving the recurrence relation from the end of the neural networks to the beginning. Moreover, we use $\sigma'$ to denote derivatives of $\sigma$. So we get
\begin{align}
&\frac{{dz_{i;\alpha }^{(L)}}}{{dz_{j;\alpha }^{(\ell )}}} = \sum\limits_{k = 1}^{{n_{\ell  + 1}}} {\frac{{dz_{i;\alpha }^{(L)}}}{{dz_{k;\alpha }^{(\ell  + 1)}}}} \frac{{dz_{k;\alpha }^{(\ell  + 1)}}}{{dz_{j;\alpha }^{(\ell )}}} = \sum\limits_{k = 1}^{{n_{\ell  + 1}}} {\frac{{dz_{i;\alpha }^{(L)}}}{{dz_{k;\alpha }^{(\ell  + 1)}}}} W_{kj}^{(\ell  + 1)}\sigma^{(\ell)} _{j;\alpha }{'}\nonumber\\
&= \sum\limits_{{k_{\ell  + 1}},{k_{\ell  + 2}}}^{{n_{\ell  + 1}},{n_{\ell  + 2}}} {\frac{{dz_{i;\alpha }^{(L)}}}{{dz_{{k_{\ell  + 2}};\alpha }^{(\ell  + 2)}}}} W_{{k_{\ell  + 2}}j}^{(\ell  + 2)}W_{{k_{\ell  + 1}}j}^{(\ell  + 1)}\sigma^{(\ell+1)} _{j;\alpha }{'}\sigma^{(L-2)} _{j;\alpha }{'}\nonumber\\
&= \sum\limits_{{k_{\ell  + 1}},{k_{\ell  + 2}}, \ldots ,{k_L}}^{{n_{\ell  + 1}},{n_{\ell  + 2}}, \ldots ,{n_L}} {\frac{{dz_{i;\alpha }^{(L)}}}{{dz_{{k_L};\alpha }^{(L)}}}} W_{{k_L}j}^{(L)}W_{{k_{L - 1}}j}^{(L - 1)} \ldots W_{{k_{\ell  + 2}}j}^{(\ell  + 2)}W_{{k_{\ell  + 1}}j}^{(\ell  + 1)}\nonumber\\
&\times \sigma^{(L-1)} _{j;\alpha }{'}\sigma^{(L-2)} _{j;\alpha }{'} \ldots \sigma^{(\ell+1)} _{j;\alpha }{'}\sigma^{(L-2)} _{j;\alpha }{'}\nonumber\\
&= \sum\limits_{{k_{\ell  + 1}},{k_{\ell  + 2}}, \ldots ,{k_{L - 1}}}^{{n_{\ell  + 1}},{n_{\ell  + 2}}, \ldots ,{n_{L - 1}}} W_{i,j}^{(L)}W_{{k_{L - 1}}j}^{(L - 1)} \ldots W_{{k_{\ell  + 2}}j}^{(\ell  + 2)}W_{{k_{\ell  + 1}}j}^{(\ell  + 1)}\nonumber\\
&\sigma^{(L-1)} _{j;\alpha }{'}\sigma^{(L-2)} _{j;\alpha }{'} \ldots \sigma^{(\ell+1)} _{j;\alpha }{'}\sigma^{(L-2)} _{j;\alpha }{'} ~.
\end{align}
We find the expectation value will vanish directly (which is exactly similar to the quantum case). Thus, we could estimate the norm by computing the variance of the gradients from,
\begin{align}\label{largewidthclassical}
&{\mathbb{E}}\left( {{{\left( {\frac{{dz_{i;\alpha }^{(L)}}}{{dz_{j;\alpha }^{(\ell )}}}} \right)}^2}} \right)\nonumber\\
&= \sum\limits_{{k_{\ell  + 1}},{k_{\ell  + 2}}, \ldots ,{k_{L - 1}},{{\bar k}_{\ell  + 1}},{{\bar k}_{\ell  + 2}}, \ldots ,{{\bar k}_{L - 1}}}^{{n_{\ell  + 1}},{n_{\ell  + 2}}, \ldots ,{n_{L - 1}},{n_{\ell  + 1}},{n_{\ell  + 2}}, \ldots ,{n_{L - 1}}} {{\mathbb{E}}\left( \begin{array}{l}
W_{i,j}^{(L)}W_{i,j}^{(L)}W_{{k_{L - 1}}j}^{(L - 1)}W_{{{\bar k}_{L - 1}}j}^{(L - 1)} \ldots\\
W_{{k_{\ell  + 2}}j}^{(\ell  + 2)}W_{{{\bar k}_{\ell  + 2}}j}^{(\ell  + 2)}W_{{k_{\ell  + 1}}j}^{(\ell  + 1)}W_{{{\bar k}_{\ell  + 1}}j}^{(\ell  + 1)}
\end{array} \right){\mathbb{E}}\left( {{{\left( {\Sigma _{j;\alpha }^{(\ell );(L - 1)}} \right)}^2}} \right)} \nonumber\\
&= \sum\limits_{{k_{\ell  + 1}},{k_{\ell  + 2}}, \ldots ,{k_{L - 1}},{{\bar k}_{\ell  + 1}},{{\bar k}_{\ell  + 2}}, \ldots ,{{\bar k}_{L - 1}}}^{{n_{\ell  + 1}},{n_{\ell  + 2}}, \ldots ,{n_{L - 1}},{n_{\ell  + 1}},{n_{\ell  + 2}}, \ldots ,{n_{L - 1}}} {{\mathbb{E}}\left( \begin{array}{l}
W_{i,j}^{(L)}W_{i,j}^{(L)}W_{{k_{L - 1}}j}^{(L - 1)}W_{{{\bar k}_{L - 1}}j}^{(L - 1)} \ldots\\
W_{{k_{\ell  + 2}}j}^{(\ell  + 2)}W_{{{\bar k}_{\ell  + 2}}j}^{(\ell  + 2)}W_{{k_{\ell  + 1}}j}^{(\ell  + 1)}W_{{{\bar k}_{\ell  + 1}}j}^{(\ell  + 1)}
\end{array} \right){\mathbb{E}}\left( {{{\left( {\Sigma _{j;\alpha }^{(\ell );(L - 1)}} \right)}^2}} \right)} \nonumber\\
&= \frac{1}{{{n_L}}}C_W^{(L)}C_W^{(L - 1)} \ldots C_W^{(\ell  + 1)}{\mathbb{E}}\left( {{{\left( {\Sigma _{j;\alpha }^{(\ell );(L - 1)}} \right)}^2}} \right)~,
\end{align}
where
\begin{align}
\Sigma _{j;\alpha }^{(\ell );(L - 1)} =\sigma^{(L-1)} _{j;\alpha }{'} \sigma^{(L-2)} _{j;\alpha }{'} \ldots \sigma^{(\ell+2)} _{j;\alpha }{'}\sigma^{(\ell+1)} _{j;\alpha }{'}~.
\end{align}
We have used the Wick contraction rule and the LeCun parametrization \ref{lecunformula} according to \cite{roberts2021principles}. Plug Equation \ref{largewidthclassical} back to Equation \ref{basicMLP}, we see that this $1/n_{L}$ factor appears. This is the classical barren plateau in the large-width classical neural networks.

\subsection{Classical large-width neural network could still learn efficiently}
Here we show that the classical neural tangent kernel (NTK) will not vanish in classical MLPs, despite its laziness. This indicates that there are many good enough local minima around the point of initialization, so even the variational angles run slowly (the barren plateau problem), it will not matter for our practical purpose. On the other hand, more variational parameters will make us converge faster. 

This part is a review of existing results, presented in the language of \cite{roberts2021principles}. In classical MLPs, similar to the quantum cases we have discussed in the whole paper, the residual training error $\varepsilon$ will decay exponentially at large width. We define the NTK as
\begin{align}
{H_{{i_1}{i_2};{\alpha _1}{\alpha _2}}} \equiv \sum\limits_\mu  {\frac{{d{z_{{i_1};{\alpha _1}}}}}{{d{\theta _\mu }}}\frac{{d{z_{{i_2};{\alpha _2}}}}}{{d{\theta _\mu }}}} ~.
\end{align}
The gradient descent rule will imply,
\begin{align}
\delta {\varepsilon _{i;\delta }} =  - \eta \sum\limits_{{i_1},\tilde \alpha  \in \mathcal{A}} {{H_{i{i_1};\delta \tilde \alpha }}{\varepsilon _{{i_1},\tilde \alpha }}}~.
\end{align}
One could compute the average of the NTK. One could define the frozen NTK and the fluctuating NTK as
\begin{align}
{H_{{i_1}{i_2};{\alpha _1}{\alpha _2}}} = {{\bar H}_{{i_1}{i_2};{\alpha _1}{\alpha _2}}} + \Delta {H_{{i_1}{i_2};{\alpha _1}{\alpha _2}}}~,
\end{align}
and we have

\begin{align}
{\mathbb{E}}\left( {\Delta {H_{{i_1}{i_2};{\alpha _1}{\alpha _2}}}\Delta {H_{{i_3}{i_4};{\alpha _3}{\alpha _4}}}} \right) = \frac{1}{{{n_{L-1}}}}\left[ {{\delta _{{i_1}{i_2}}}{\delta _{{i_3}{i_4}}}{A_{\left( {{\alpha _1}{\alpha _2}} \right)\left( {{\alpha _3}{\alpha _4}} \right)}} + {\delta _{{i_1}{i_3}}}{\delta _{{i_2}{i_4}}}{B_{{\alpha _1}{\alpha _3}{\alpha _2}{\alpha _4}}} + {\delta _{{i_1}{i_4}}}{\delta _{{i_2}{i_3}}}{B_{{\alpha _1}{\alpha _4}{\alpha _2}{\alpha _3}}}} \right]~.
\end{align}

The full expressions of $A,B$ are given in Chapter 8 of \cite{roberts2021principles}. Similarly, in the statistics language, one could check \cite{jacot2018neural}. The suppression of $\Delta H$ in the large width indicates that the large-width neural networks will learn efficiently through non-trivial $ {{\bar H}_{{i_1}{i_2};{\alpha _1}{\alpha _2}}} $, which is guaranteed to converge exponentially. In the large-width limit, the gradient descent algorithm is theoretically equivalent to the kernel method, where the kernel is defined effectively by NTKs. In Chapter 11 of \cite{roberts2021principles}, it is shown that dNTK, the higher-order corrections to the exponential decay, will vanish on its own, averaging over the Gaussian distribution of weights and bias. Moreover, the correlations between dNTK and other operators, which cause even numbers of $W$s in total, will be suppressed by the large width polynomially. Those theoretical results are classical analogs of random unitary calculations done in our work.

\section{Some further details about concentration conditions}\label{concenreview}
For concentration conditions including the quantum meta-kernel, one could see \cite{together} for further details. Here we provide a simple review. 

Now, we would like to ask when the QNTK approximation is valid. When the learning rate is small, the error of the prediction in Equation \ref{fullqntke} could possibly come from two sources: the fluctuation of $K$ about $\bar{K}$ during the gradient descent, and the higher-order corrections comparing the leading order Taylor expansion in Equation \ref{taylor}. The fluctuation $\Delta K$ could come from higher-order statistical calculations over the $k$-design assumption, similar to the analysis of higher-order effects in the barren plateau setup \cite{cerezo2021higher}, 
\begin{align}
\Delta K = \sqrt {{\mathbb{E}}\left( {{{(K - \bar K)}^2}} \right)}  \approx \frac{{\sqrt L }}{{{N^2}}}\sqrt {\left( {8{{{\mathop{\rm Tr}\nolimits} }^2}\left( {{O^2}} \right) + 12{\mathop{\rm Tr}\nolimits} \left( {{O^4}} \right)} \right)} ~,
\end{align}
in the large-$N$ limit, and we present a detailed calculation in \cite{together} with formulas up to 4-design. Moreover, we could look at higher order corrections to the Taylor expansion by the quantum meta-kernel (dQNTK) \cite{Liu:2021wqr},
\begin{align}
&\delta \varepsilon  =  - \eta \sum\limits_\ell  {\frac{{d\varepsilon }}{{d{\theta _\ell }}}} \frac{{d\varepsilon }}{{d{\theta _\ell }}}\varepsilon  + \frac{1}{2}{\eta ^2}{\varepsilon ^2}\sum\limits_{{\ell _1},{\ell _2}} {\frac{{{d^2}\varepsilon }}{{d{\theta _{{\ell _1}}}d{\theta _{{\ell _2}}}}}} \frac{{d\varepsilon }}{{d{\theta _{{\ell _1}}}}}\frac{{d\varepsilon }}{{d{\theta _{{\ell _2}}}}}\nonumber\\
&\equiv  - \eta K\varepsilon  + \frac{1}{2}{\eta ^2}{\varepsilon ^2}\mu ~.
\end{align}
Here $\mu=\sum_{\ell_{1}, \ell_{2}} \frac{d^{2} \varepsilon}{d \theta_{\ell_{1}} d \theta_{\ell_{2}}} \frac{d \varepsilon}{d \theta_{\ell_{1}}} \frac{d \varepsilon}{d \theta_{\ell_{2}}}$ could be computed statistically using $k$-design formulas again. One can show that $\mathbb{E}(\mu) =0 $ (which is the same as its classical counterpart \cite{roberts2021principles}), and we have
\begin{align}
\Delta \mu  = \sqrt {{\mathbb{E}}\left( {{\mu ^2}} \right)}  \approx \frac{{\sqrt {32} L}}{{{N^3}}}{{\mathop{\rm Tr}\nolimits} ^{3/2}}\left( {{O^2}} \right)~,
\end{align}
in the large-$N$ limit. The condition where the QNTK estimation in Equation \ref{fullqntke} is valid when 
\begin{align}
&\Delta K \ll K \Leftrightarrow L \gg 1 ~,\label{cond1}\\
&\frac{1}{2}{\eta ^2}{\varepsilon ^2}\Delta \mu  \ll \eta \bar K\varepsilon  \Leftrightarrow \eta {\varepsilon(0)}\frac{L}{{{N^3}}}{{\mathop{\rm Tr}\nolimits} ^{3/2}}\left( {{O^2}} \right) \ll \frac{{L{\mathop{\rm Tr}\nolimits} \left( {{O^2}} \right)}}{{{N^2}}}  \nonumber\\
&\Leftrightarrow \frac{{\eta {\Omega _O}}}{N}{\varepsilon(0)} \ll 1~.\label{cond2}
\end{align} 
We call the conditions \ref{cond1} and \ref{cond2} as the \emph{concentration conditions}. Here, we denote $\varepsilon(0)=\varepsilon(t=0)$, and we assume that $\text{Tr}(O^2)\equiv \Omega_O^2 > \text{Tr}^2(O)$. This is correct, for instance, if $O$ is a Pauli operator, where we have $\text{Tr}(O^2)=N$ but $\text{Tr}^2(O)=0$.

Note that the condition Equation \ref{cond2} is a weak condition. It only tells that how small $\eta$ is needed to make sure the nearly expansion is valid. In practice, we often assume that $\eta < \mathcal{O}(1)$ and $\Omega_O \ge  \mathcal{O} (N)$, so Equation \ref{cond2} is automatically satisfied. The condition that usually matters is Equation \ref{cond1}, which is the definition of overparametrization here $L\gg 1$. Thus, if $L$ is large, the prediction will be correct, no matter how large $N$ is. But if $N$ is large, the decay rate itself $\bar{K}$ will be small. So this is exactly the definition of the barren plateau! 

Furthermore, we wish to mention that if we only count for powers of $N$ and $L$, we have
\begin{align}
\frac{{\Delta K}}{{\bar K}} = \mathcal{O}\left( {\frac{1}{{\sqrt L }}} \right)~,\frac{{\Delta \mu }}{{\bar K}} = \mathcal{O}\left( {\frac{1}{N}} \right)~.
\end{align}
If we demand $\bar{K}=\mathcal{O}(1)$ and ignore $\eta$, we get $L= \mathcal{O}(N)$, so we get $\frac{{\Delta K}}{{\bar K}} = \mathcal{O}\left( {\frac{1}{{N }}} \right)$ as well. The $1/N$ or $1/\text{width}$ expansion is exactly observed in the classical neural networks \cite{roberts2021principles}. The origin of this equivalence comes from the similarity between Equation \ref{1design} and Equation \ref{lecun}, while a higher level (but heuristic) understanding comes from a connection between quantum field theory and the large-width expansion \cite{dyer2019asymptotics,halverson2021neural,roberts2021principles} and a similarity between Feynman rules in quantum field theory and matrix models \cite{Witten:1995ex}, which we will briefly explain in Appendix \ref{stringtheory} for readers who are interested in how observations about this paper might be discovered from another perspective. 

\section{A physical interpretation}\label{stringtheory}

Here we make some comments about possible, heuristic, physical interpretations of the agreement between classical and quantum neural networks. There is a duality, pointed out in \cite{dyer2019asymptotics,halverson2021neural,roberts2021ai,roberts2021principles} where the large-width classical neural networks could be understood in the quantum field theory language. In the large-width limit, the output of neural networks will follow a Gaussian process, averaging with respect to Gaussian distribution over weights and bias according to the LeCun parametrization, 
\begin{align}\label{lecun}
\mathbb{E}\left(W_{i j} W_{k l}\right)=\frac{\sigma_W^2}{\text { width }} \delta_{i k} \delta_{j l}~,
\end{align}
or more generally, 
\begin{align}
\mathbb{E}\left( {{W_{{i_1}{j_1}}}{W_{{i_2}{j_2}}} \ldots {W_{{i_{2k - 1}}{j_{2k - 1}}}}{W_{{i_{2k}}{j_{2k}}}}} \right) = \mathcal{O}(\frac{1}{{{\mathop{\rm poly}\nolimits} ({\rm{width}})}})~,
\end{align}
for all positive integer $k$. Here, we are considering the multilayer perceptron (MLP) model with weights $W$, and the width is defined as the number of neurons in each layer. The limit is mathematically similar to the large-$N$ limit of gauge theories, which becomes almost generalized free theories. We could understand the ratio between the depth, the number of layers, and the width, the number of neurons, as perturbative corrections against the Gaussian process, which is similar to what we have done in the large-$N$ expansion of gauge theories. 

This physical interpretation will be helpful also when we consider its quantum generalization. If classical MLPs are similar to quantum field theories, quantum neural networks will be similar to matrix models \cite{Banks:1996vh,Berenstein:2002jq}. Matrix models have been studied for a long time, around and after the second string theory revolution \cite{Witten:1995ex}, and they have deep connections to the holographic principle \cite{Susskind:1994vu} and the AdS/CFT correspondence \cite{Maldacena:1997re,Witten:1998qj}. Haar ensembles are toy versions of matrix models, which have been widely studied as toy models of chaotic quantum black holes \cite{Hayden:2007cs,Roberts:2016hpo}. The similarity between the LeCun parametrization \ref{lecun} and the 1-design Haar integral formula
\begin{align}
\mathbb{E}({U_{ij}}U_{kl}^\dag ) = \frac{1}{{\dim \mathcal{H}}}{\delta _{il}}{\delta _{jk}}~,
\end{align}
or more generally,
\begin{align}
\mathbb{E}\left( {{U_{{i_1}{j_1}}}U_{{i_2}{j_2}}^\dag  \ldots {U_{{i_{2k - 1}}{j_{2k - 1}}}}U_{{i_{2k}}{j_{2k}}}^\dag } \right) = \mathcal{O}(\frac{1}{{{\mathop{\rm poly}\nolimits} (\dim \mathcal{H})}})~,
\end{align}
where $\dim \mathcal{H}$ is the dimension of the Hilbert space, might be potentially related to the similarity of Feynman rules between matrix models and quantum field theories. Thus, the similarity between quantum and classical neural networks might have a physical interpretation between matrix models and their effective field theory descriptions. 

The above analogy is heuristic. We should point out that machine learning and physical systems are very different. Some mathematical similarities could provide guidance towards new discoveries and better insights, but we have to be careful that they are intrinsically different phenomena.

\section{Noises}\label{noise}
Now let us add the affection of the noise. From the original gradient descent equation,
\begin{align}
{\theta _\ell }(t + 1) - {\theta _\ell }(t) \equiv \delta {\theta _\mu } =  - \eta \frac{{\partial {\cal L}}}{{\partial {\theta _\ell }}} = i\eta \left\langle {{\Psi _0}\left| {V_{ + ,\ell }^\dag \left[ {{X_\ell },V_{ - ,\ell }^\dag O{V_{ - ,\ell }}} \right]{V_{ + ,\ell }}} \right|{\Psi _0}} \right\rangle ~,
\end{align}
we add a random fluctuation term $\Delta \theta_\ell$ to model the uncertainty of measuring the expectation value. We assume that the random variable $\Delta \theta_\ell$ is Markovian. Namely, it is independent for the time step $t$. Moreover, we assume that $\Delta \theta _\ell $s are distributed with Gaussian distributions $\mathcal{N}(0,\sigma^2_\theta)$. 

Thus, the residual training error has the recursion relation in the linear order of the Taylor expansion,
\begin{align}
\delta \varepsilon =  - \eta \varepsilon K + \sum\limits_\ell  {\frac{{\partial \varepsilon }}{{\partial {\theta _\ell }}}\Delta \theta _\ell } ~.
\end{align}
Now, let us assume that $K$ is still a constant. Since $\Delta \theta_\ell \sim \mathcal{N}(0,\sigma^2_\theta)$, we get
\begin{align}
\sum\limits_\ell  {\frac{{\partial \varepsilon }}{{\partial {\theta _\ell }}}\Delta {\theta _\ell }}  \sim \mathcal{N}(0,K\sigma _\theta ^2)~.
\end{align}
Thus, we could write the recursion relation as
\begin{align}
\delta \varepsilon =  - \eta \varepsilon K + \sqrt{K} \Delta \theta ~.
\end{align}
Here, $\Delta \theta \approx \mathcal{N} (0, \sigma _\theta^2)$. One can solve the difference equation iteratively. The answer is
\begin{align}
\varepsilon (t) = {(1 - \eta K)^t}\varepsilon (0) + \sqrt K \sum\limits_{i = 0}^{t - 1} {{{(1 - \eta K)}^i}\Delta \theta (t - 1 - i)} ~.
\end{align}
Now, we have
\begin{align}
&\sqrt K \sum\limits_{i = 0}^{t - 1} {{{(1 - \eta K)}^i}\Delta \theta (t - 1 - i)}  \sim {\cal N}(0,K\sigma_\theta^2\sum\limits_{i = 0}^{t - 1} {{{(1 - \eta K)}^{2i}}} )\nonumber\\
&= {\cal N}(0,\sigma_\theta^2 \frac{{1 - {{(1 - \eta K)}^{2t}}}}{{\eta (2 - \eta K)}})~.
\end{align}

At the initial time $t=0$, there is no effect of noise. The relative size of the error will grow during time compared to the exponential decay term without noises. Based on the distribution, we could compute the average $\varepsilon^2$ against the noises, $\overline {{\varepsilon ^2}} $, as 
\begin{align}
\overline {{\varepsilon ^2}} (t)= {(1 - \eta K)^{2t}}\left( {{\varepsilon ^2}(0) - \frac{{\sigma _\theta ^2}}{{\eta (2 - \eta K)}}} \right) + \frac{{\sigma _\theta ^2}}{{\eta (2 - \eta K)}}~.
\end{align}
Note that the first term is decaying when the time $t$ is increasing. At the late time, we have
\begin{align}
\overline {{\varepsilon ^2}} (\infty ) = \frac{{\sigma _\theta ^2}}{{\eta (2 - \eta K)}} \approx \mathcal{O}(\frac{{\sigma _\theta ^2}}{\eta })~,
\end{align}
where we assume the overparametrization $\eta K \approx \mathcal{O}(1)$~. Thus, at the late time, the loss function will arrive at a constant plateau at $\mathcal{O}({{\sigma _\theta ^2}}/{\eta })$. One could improve $\sigma_\theta$ to make the constant plateau controllable and do not increase significantly with $N$, indicating that our algorithm could be noise-resilient. 

One could also estimate the time scale where the contribution of the noise could emerge. We could define the time scale, $T_\text{noise}$, as,
\begin{align}
{(1 - \eta K)^{{T_{{\rm{noise}}}}}}\varepsilon (0) \approx {\sigma _\theta }\sqrt {\frac{{1 - {{(1 - \eta K)}^{2{T_{{\rm{noise}}}}}}}}{{\eta (2 - \eta K)}}} ~.
\end{align}
It means that at $T_{\text{noise}}$, the noise contribution is comparable to the noiseless part in the residual training error. We have,
\begin{align}\label{tnoise}
&{T_{{\rm{noise}}}} \approx \frac{{\log \left( {\frac{{{\sigma _\theta }}}{{\sqrt {2{\varepsilon ^2}(0)\eta  - {\varepsilon ^2}(0){\eta ^2}K + \sigma _\theta ^2} }}} \right)}}{{\log (1 - \eta K)}}~,\nonumber\\
 &\varepsilon (T_{\rm{noise}})=2{(1 - \eta K)^{{T_{{\rm{noise}}}}}}\varepsilon (0)=\frac{2\sigma_\theta^2}{\sqrt{\varepsilon (0)^2(2\eta-\eta^2K)+\sigma_\theta^2}}\varepsilon (0)~.
\end{align}
We find that choosing $\eta \approx \mathcal{O}(1/K)$ will minimize $\varepsilon (T_{\rm{noise}})$. It is exactly the overparametrization condition we use in this paper. 

To be self-consistent, we need to check if the choice $\eta \approx \mathcal{O}(1/K)$ is consistent with the concentration condition about dQNTK. In fact, we find that $\eta \approx \mathcal{O}(1/K)$ will naturally satisfy the dQNTK concentration condition if $ \varepsilon(0)<\mathcal{O}(L\sqrt{N})$. This is naturally satisfied in generic situations in variational quantum algorithms since we will usually not have an exponential amount of residual training error initially.

\section{Numerical results}\label{numerical}
\begin{figure}
    \centering
    \includegraphics[width=13cm]{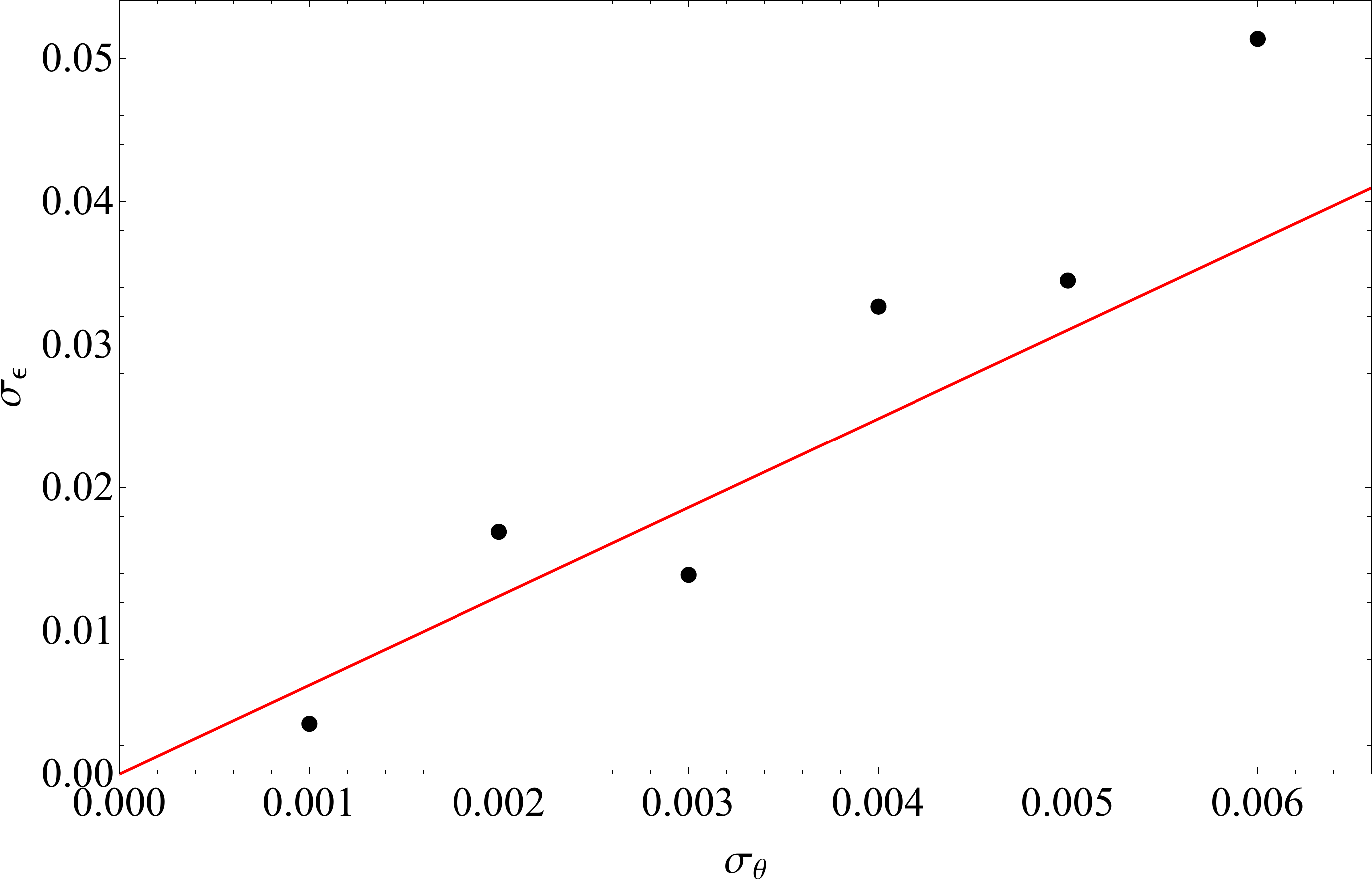}
    \caption{Noise standard deviation $\sigma_\theta$ as a function of standard deviation of final residual error $\sigma_\varepsilon$ after training long enough time, with both numerical result (black dots) and theoretical prediction (red line). In this figure, $\eta=0.005$, $K\approx 25$, $\varepsilon(0)\approx 1$.}
    \label{fignew}
\end{figure}

In this part, we show some simple numerical evidences based on the analysis done in \cite{together}. We will use the randomized version of the hardware-efficient variational ansatz defined in \cite{together}. In Figure \ref{fignew}, for each $\sigma_\theta$ value, we run 10 experiments of 100 steps using the same setup of the ansatz $U(\theta)$, the operator $O$ and the input state $\theta_0$ as in \cite{together}. After that, we get the residual error of the last step and take the average value over 10 experiments to get the mean $\varepsilon$ value, shown with black dots in the figure. The red line in the figure is the theoretical prediction. In these experiments, $L=64$, and we have 4 qubits. We can further get the analytic result of the mean value of $\overline{\varepsilon}$ after a long time as
\begin{equation}
    \overline{\varepsilon}=\sqrt{\frac{2}{\pi}}\cdot\frac{\sigma_\theta}{\sqrt{2\eta-\eta^2K}}~,
\end{equation}
where the $K$ value is taken from the value of the last step, as it fluctuates a lot in the early time. 

We run multiple experiments to approach the theoretical value as much as possible, where 10 experiments are done for each $\sigma_\theta$ value. To verify that the numerical result lies in a reasonable regime, we calculated the 90\% confidence interval of $\varepsilon$ theoretically.

To compensate for the effect of large $K$ on our numerical simulations, since in every experiment setup, due to randomness, the training will lead the parameters to different regimes of different $K$s, we choose those experiments which fulfill our theoretical restrictions for small $K$. The numerical results above are with $K\approx\mathcal{O}(10)$, which still shows great agreement with our theoretical formalism.

More precisely, in Figure \ref{fignew}, we get the relationship between residual error fluctuation and noise. For each $\sigma_\theta$ value, we calculated the standard deviation with final residual error data from 10 experiments, shown as black dots. The final residual error that we get from the numerical experiments is taken absolute value for the benefit of the log scale. We find the numerical results follow the theoretical prediction in a reasonable confidence interval. Moreover, we verify the extent of our final residual error that can achieve as a function of noise $\sigma_\theta$ with numerical evidence.

\begin{figure}
    \centering
    \includegraphics[width=13cm]{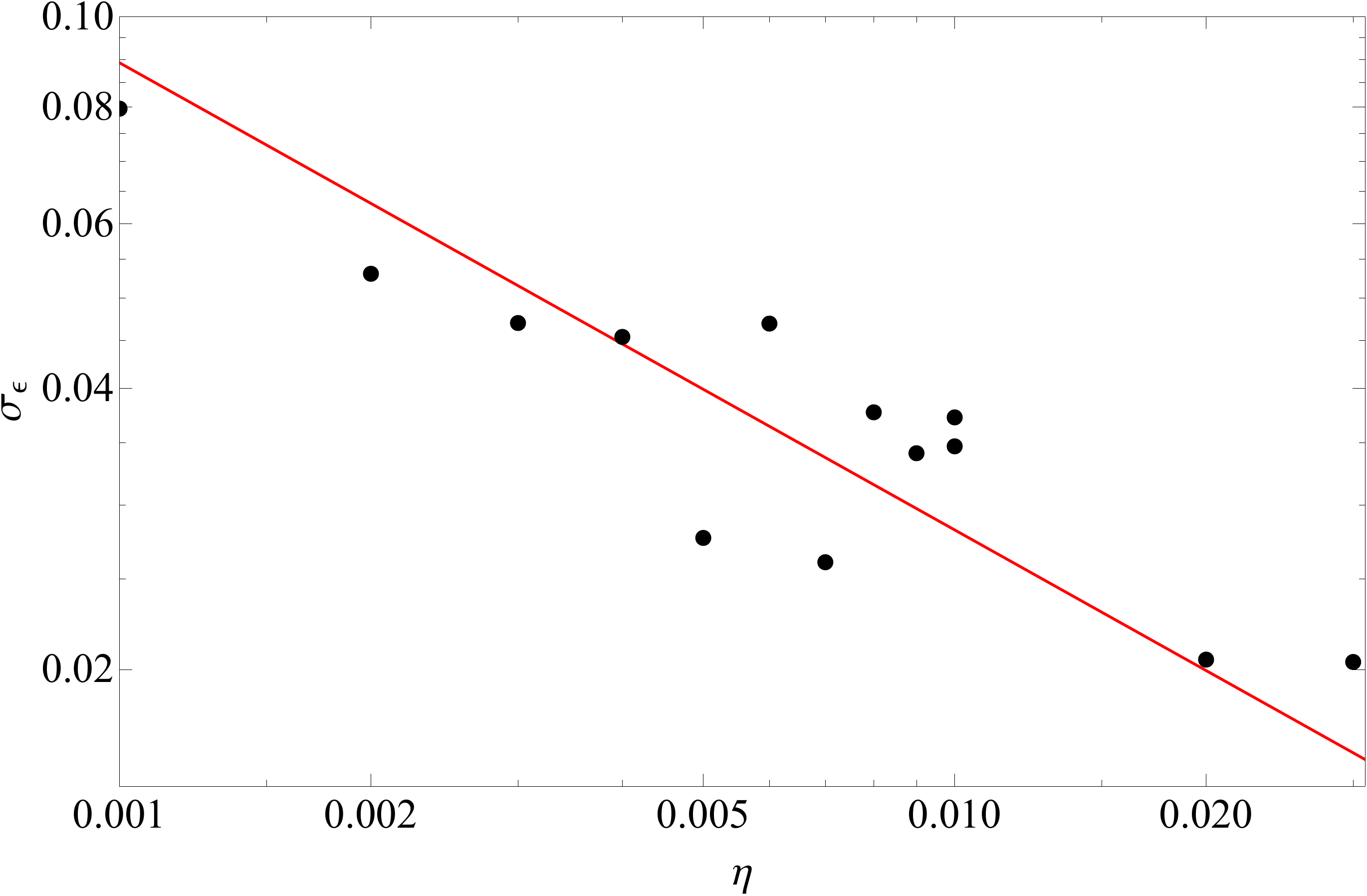}
    \caption{Standard deviation of final residual error $\sigma_\varepsilon$ as a function of learning rate $\eta$ after training long enough time, with both numerical result (black dots) and theoretical prediction (red line). In this figure, $\sigma_\theta=0.005$, $K\approx 35$, $\varepsilon(0)\approx 1$, $t=100$.}
    \label{sigmathetavseta}
\end{figure}

In Figure \ref{sigmathetavseta}, we verify the prediction of standard deviation of $\varepsilon(\infty)$, $\sigma_\varepsilon$, in the small $\eta$ regime. In these numerical experiments, the inaccuracy comes mainly from a limited number of experiments and a limited time scale ($t=100$). Especially for experiments with a small learning rate $\eta$ with random initial states, $T_{\rm{noise}}$ may be large for 100 steps to cover.

\bibliography{ref.bib}

\end{document}